\pdfoutput=1

\documentclass[11pt]{article}

\usepackage[]{ACL2023}

\usepackage{times}
\usepackage{latexsym}
\usepackage{graphicx}
\usepackage{multirow}
\usepackage{url}
\usepackage[]{fancyhdr}
\usepackage{booktabs}
\usepackage[T1]{fontenc}
\usepackage{tipa}

\usepackage{tikz}
\def\checkmark{\tikz\fill[scale=0.4](0,.35) -- (.25,0) -- (1,.7) -- (.25,.15) -- cycle;} 

\usepackage[utf8]{inputenc}

\usepackage{microtype}
\usepackage[ethiop,main=english]{babel}
\usepackage{inconsolata}

\title{A Case Against Implicit Standards: Homophone Normalization in Machine Translation for Languages that use the Ge'ez Script}

\fancypagestyle{plain}{%
  \fancyhf{}
\lhead{ Preprint. Under review.
}

}

\author{
 \textbf{Hellina Hailu Nigatu\textsuperscript{1}},
 \textbf{Atnafu Lambebo Tonja\textsuperscript{2}},
 \textbf{Henok Biadglign Ademtew\textsuperscript{3}},\\
 \textbf{Hizkel Mitiku Alemayehu\textsuperscript{4}},
 \textbf{Negasi Haile Abadi\textsuperscript{5}},
 \textbf{Tadesse Destaw Belay\textsuperscript{6}},
 \textbf{Seid Muhie Yimam\textsuperscript{7}},
\\
\\
 \textsuperscript{1}UC Berkeley,
 \textsuperscript{2} MBZUAI,
 \textsuperscript{3} Vella AI,
 \textsuperscript{4} Paderborn University,
 \textsuperscript{5} Lesan AI,\\
 \textsuperscript{6}Instituto Politécnico Nacional,
 \textsuperscript{7}University of Hamburg
\\
 \small{
   \textbf{Correspondence:} \href{mailto:hellina\_nigatu@berkeley.edu}{hellina\_nigatu@berkeley.edu}
 }
}

\begin{document}
\pagestyle{fancy}
\fancyhead{} 
\fancyhead[L]{\textbf{Preprint. Under Review.}}

\maketitle
\thispagestyle{fancy}

\begin{abstract}

Homophone\footnote{We use the \citet{merriam-webster_definition_nodate} definition of the term homophone: ``a character or group of characters pronounced the same as another character or group''.} normalization--where characters that have the same sound in a writing script are mapped to one character--is a pre-processing step applied in Amharic Natural Language Processing (NLP) literature. While this may improve performance reported by automatic metrics, it also results in models that are not able to understand different forms of writing in a single language. Further, there might be impacts in transfer learning, where models trained on normalized data do not generalize well to other languages. 
In this paper, we experiment with monolingual training and cross-lingual transfer to understand the impacts of normalization on languages that use the Ge'ez script. 
We then propose a post-inference intervention in which normalization is applied to model predictions instead of training data.
With our simple scheme of post-inference normalization, we show that we can achieve an increase in BLEU score of up to 1.03 while preserving language features in training. 
Our work contributes to the broader discussion on technology-facilitated language change and calls for more language-aware interventions.

\end{abstract}

\section{Introduction}


The majority of the world's languages are underrepresented in natural language processing (NLP) research~\cite{joshi_state_2020}. 
Collectively, these languages have been referred to as `low-resource,' owing to the various resources that are not available for them \cite{nigatu_zenos_2024}. 
One of the many resources that are lacking for low-resourced languages is pre-processing tools~\cite{niyongabo-etal-2020-kinnews}. From tokenization methods to basic data cleaning tools, many of the widely used pre-processing schemes do not include, or are not effective for, low-resourced languages~\cite{ahia_all_2023, emezue_african_2023}. 


Pre-processing steps, ranging from removing punctuation marks to tokenizing text, are essential steps in determining the efficacy of downstream models.
For instance, languages that use different writing scripts have been transliterated to a single script to facilitate cross-lingual transfer~\cite{khare_low_2021}.
Looking at tokenization, morpheme-based tokenization has been used for morphologically rich languages in place of word-level tokenization to improve performance \cite{tachbelie_using_2014}. 
Within phonetic languages like Amharic, a common pre-processing intervention has been homophone normalization--i.e, mapping characters with similar sounds to one character~\cite{biadgligne_parallel_2021, abate_parallel_2018}. 

A few prior works have explored the impacts of homophone normalization. \citet{abate_multilingual_2020} find that homophone normalization reduces the vocabulary size of the dataset by 16\%, which may be desirable for some applications. 
\citet{belay_effect_2022} compared machine translation models with and without normalization and found that normalization boosts the BLEU score by 2.27 percentage points for English-Amharic translation with a Transformer model trained from scratch. 
While this indicates a potential benefit in improving performance when using automatic metrics for evaluation, it may lead to downstream issues for language users. 


In this paper, we argue that the seemingly innocuous act of homophone normalization for Amharic NLP sets an implicit standard for Ge'ez script languages.  
First, models trained on normalized datasets will not be able to understand alternative word spellings--limiting how users can interact with language technologies in their language. 
Second, homophone normalization within NLP literature is applied to one of the many languages that use the Ge'ez script. In other languages, such as Tigrinya and Ge'ez, the characters that are normalized have distinct sounds.
Hence, the implicit standard set by this pre-processing step may have a downstream impact on cross-lingual transfer for the other languages that use the Ge'ez script.
Using Machine Translation (MT) as an NLP task and Amharic, Tigrinya, and Ge'ez as languages of focus, we pose the following research questions:


\begin{itemize}
    \item \textbf{RQ1:} How do existing MT models handle words with homophone characters in languages that use the Ge'ez script?

    \item \textbf{RQ2:} What is the impact of applying different normalization schemes to training data on the performance of MT systems? 
    
    \item \textbf{RQ3:}  What is the impact of homophone normalization on transfer learning in MT for related languages?  

    \item \textbf{RQ4:} How does applying normalization post-translation compare to applying normalization to the training data?

\end{itemize}

Multilingual NLP research is often driven by a goal of generalization, proposing ways to make a single model work well for multiple languages~\cite[e.g][]{nllb_no_2022}. 
While there are demonstrated benefits to this approach, we use our work as a case study to question what we lose through implicit standards in language processing. 
We find that homophone normalization negatively affects cross-lingual transfer and that applying normalization post-translation boosts automatic scores without compromising language characteristics (Sec. \ref{sec:experiments}). Our work highlights the importance of investigating downstream impacts of pre-processing steps, particularly for low-resourced languages\footnote{Models, code and data can be found at \url{https://github.com/hhnigatu/geez\_script\_normalization}}. 




\section{Background and Related Work}\label{sec:background}
In this section, we provide background on the languages of study and the writing script. We also give background on normalization schemes used in prior work to handle characters with the same sound.  

\subsection{Languages of Study}
\paragraph{The Ge'ez Script} is an Abugida writing system--each character in the script represents a consonant and a vowel\footnote{https://www.omniglot.com/writing/ethiopic.htm}. Vowels are indicated by modifying the base character, which is the consonant. There are 7 vowels in the Ge'ez writing script; hence, each base character has at least 7 variations. For instance, the 
base character `\selectlanguage{ethiop} la' \selectlanguage{english} /l/ is used to represent the sound /la/ and is modified to `\selectlanguage{ethiop} lu' \selectlanguage{english} /lu/, `\selectlanguage{ethiop} li' \selectlanguage{english}/li/, and so on. Additionally, there are characters used to represent labiovelars such as `\selectlanguage{ethiop} kuA'\selectlanguage{english} /kwa/.  The Ge'ez script is used to write Afro-Semitic languages of Ethiopia and Eritrea, including our languages of focus in this paper: Amharic, Tigrinya, and Ge'ez.

\paragraph{Amharic} is an Afro-Semitic language spoken by an estimated 57.5 million people worldwide~\cite{basha_detection_2023}. It is primarily spoken in Ethiopia and is one of the federal working languages of the country. The Amharic alphabet has 33 base characters~\cite{adugna_research_nodate}. 

\paragraph{Tigrinya} is an Afro-Semitic language spoken by an estimated 10 million people worldwide~\cite{haile_error_2023}. 
Tigrinya is one of the federal working languages of Ethiopia and is one of the governmental and national languages of Eritrea. The Tigrinya alphabet has 32 base characters~\cite{negash_origin_2017}. 

\paragraph{Ge'ez} is an Afro-Semitic language that is labeled as extinct\footnote{https://www.ethnologue.com/language/gez/} as it has no native speakers. It is primarily used as a liturgical language within Ethiopian and Eritrean religious institutions. 
The Ge'ez alphabet has 26 base characters \cite{demilew_ancient_2019}. 

\subsection{Homophones in the Ge'ez Script}
As languages evolve, phonological change occurs where some phonemes might split, merge, or emerge~\cite{boldsen-paggio-2022-letters}. Since written language evolves at a much slower pace than spoken language, the phonetic changes are usually not reflected in the written forms of language~\cite{obasi_structural_nodate}.  
Due to merged phonemes that are represented by different characters that, in prior years, might have had distinct sounds, the Amharic alphabet has multiple characters that have the same sound~\cite{aklilu_problems_2010}. 
For instance, the sound /\textglotstop/ can be written by any of the following characters in the Ge'ez script: `\selectlanguage{ethiop}'a\selectlanguage{english}', `\selectlanguage{ethiop}'A'\selectlanguage{english}, \selectlanguage{ethiop}‘a' \selectlanguage{english} or \selectlanguage{ethiop}‘A`. \selectlanguage{english}

Writing scripts are also shared by several languages that may not have evolved in the same way. For the Ge'ez script in particular, some of the characters that have the same sound in Amharic have distinct sounds in Tigrinya. For example, all four characters in the above example that represent the /\textglotstop/ sound in Amharic have distinct sounds in Tigrinya: `\selectlanguage{ethiop}'a\selectlanguage{english}' /\textglotstop \textturna/, `\selectlanguage{ethiop}'A\selectlanguage{english}' /\textglotstop \"a/, \selectlanguage{ethiop}‘a\selectlanguage{english}' /\textrevglotstop \textturna/ and \selectlanguage{ethiop}‘A\selectlanguage{english}/\textrevglotstop\"a/.  
There are some characters from the Ge'ez script that have the same sound in Tigrinya, for example `\selectlanguage{ethiop}'sa\selectlanguage{english}' and `\selectlanguage{ethiop}sa'\selectlanguage{english} both represent the sound /s\textturna/.  
Due to the differences in how each language uses the characters, altering homophones in the Ge'ez script will have different effects across languages. 
For instance, if you write the word \selectlanguage{ethiop}‘Ayene'\selectlanguage{english} in Tigrinya as `\selectlanguage{ethiop}'ayene'\selectlanguage{english}, the word would have no meaning, while in Amharic, both words would mean `eye'.
In the Ge'ez language, changing the characters will result in a change in meaning. For instance, the word `\selectlanguage{ethiop}saraga' \selectlanguage{english} means `to hold a wedding' while the word `\selectlanguage{ethiop}'saraga'\selectlanguage{english} means `to get inside'.


\subsection{Handling Homophone Characters in NLP}

Current NLP evaluation schemes, particularly automatic metrics like BLEU~\cite{papineni-etal-2002-bleu}, which require an exact match between n-grams, do not handle the Amharic homophone characters. For instance, if the Amharic word \selectlanguage{ethiop}‘Ayene' \selectlanguage{english}is written as `\selectlanguage{ethiop}'ayene\selectlanguage{english}' in the reference but the model prediction outputs \selectlanguage{ethiop}‘Ayene'\selectlanguage{english}, the evaluation would not count it as a match even though, for a person who speaks the Amharic language, those two words have the same pronunciation and meaning. 

Homophone normalization averts this problem by mapping all homophone characters into a single character. It has mainly been applied in Amharic NLP literature for Machine Translation~\cite[e.g.][]{abate_parallel_2018, chekole_effect_2024} and semantic modeling tasks~\cite[e.g.][]{belay_impacts_2021}. 
However, within papers that report normalizing homophone characters, there is no standard normalization scheme. For instance, some publicly available tools normalize characters with the same sound only~\cite[e.g.][]{kidanemariam_amharic-nlp-tools--java_2019}, others normalize characters with the same sound and labialized characters~\cite{mekuriaw_benchmark_2024, eshetu_amharic-simple-text-preprocessing-usin-python_2022}, and some normalize characters with the same sound, labialized characters, and some characters with the same base consonant~\cite{yimam_introducing_2021}. Further, some prior works report mapping homophone characters to `the most frequently used characters''~\cite{biadgligne_offline_2022, abate_parallel_2018}. 



While most prior works report using normalization as a standard pre-processing step, 
\citet{belay_effect_2022} compared MT models trained with and without normalization and reported score improvement for models trained with normalized data. 
\citet{belay_impacts_2021} applies normalization to semantic modeling tasks and finds that normalization helps for Information Retrieval but hurts performance for PoS tagging and sentiment analysis. 
However, these investigations are (1) limited to the Amharic language and (2) do not compare the impact of the different normalization schemes in the literature.


\paragraph{Cases for and against homophone normalization in Amharic:}
From linguistics literature, there have been three viewpoints on how to handle characters that have the same sound in Amharic: (1) standardize spellings, (2) remove homophone characters from the alphabet--i.e, normalize, or (3) perform no interventions~\cite{aklilu_problems_2010}.

Thus far, the Amharic NLP literature has adopted an (implicit) standardization step with homophone normalization. 
In this paper, we offer a post-inference intervention that provides a middle ground to the three viewpoints described above. Instead of training on normalized data, we propose performing normalization when calculating a particular metric. We first investigate the impacts of normalization in MT in zero-shot, monolingual, and cross-lingual settings and show that our post-inference intervention can improve metric scores.




\section{Methods} \label{sec:methods}
To test the impact of homophone normalization, we prepared an evaluation dataset with a focus on words that have homophone characters in the three languages using publicly available MT datasets (Sec. \ref{sec:dataset}). We then adopted two normalization schemes for our experiments, which we describe in Sec. \ref{sec:normalization}.

\subsection{Dataset} \label{sec:dataset}
We prepared an evaluation dataset in the three languages of study by focusing on sentences that have high counts of characters with the same sound. We selected sentences from the following datasets for each language:

\paragraph{Amharic-English-Machine-Translation-Corpus} The Amharic-English Machine Translation Corpus \cite{abate_parallel_2018} contains Amharic-English parallel sentences collected from Bible, History, News, and Legal sources. The dataset has a total of 276k parallel sentences. From the test split of the \cite{abate_parallel_2018} dataset, we selected sentences that had at least 9 homophone characters. With this filtering step, we had a test set of 2.4k 
Amharic-English sentence pairs.

\paragraph{Tigrinya-English MT} For Tigrinya, we used data from \citet{lakew_low_2020} and \citet{abate_parallel_2018}. The dataset had a total of 150.8k parallel sentences. Similar to Amharic, we selected sentences that had at least 17 homophone characters, which resulted in a test set with 2.4k English-Tigrinya parallel sentences. 

\paragraph{AGE} We used the AGE dataset~\cite{ademtew_age_2024} which has 17.5k Amharic-Ge'ez and 18.6k Ge'ez-English parallel sentences. 
For our experiments, we used the English-Ge'ez data and split it into training, evaluation, and test sets at an 8:1:1 ratio. With this, we had 1964 Ge'ez-English parallel sentences as our test set. Since the Ge'ez dataset is small, we did not apply additional filtering to the test set. 

Our final evaluation dataset has 2.4k sentences for Tigrinya and Amharic and 1.9k sentences for Ge'ez. In Table \ref{tab:dataset}, we give the details of our dataset\footnote{We will add a link to data, code, and models upon publication}. 

\begin{table}[]
    \centering
    \small
   \begin{tabular}{c|p{2cm}|c|c|c}
   \toprule
      \textbf{Language} & \textbf{Source Dataset} & \textbf{Training} & \textbf{Eval} & \textbf{Test}  \\
      \midrule
       Amharic &\citet{abate_parallel_2018} &199.2k &22.1k & 2.4k\\ \hline
       Ge'ez & \citet{ademtew_age_2024} & 15.7k & 1.9k & 1.9k \\
       \hline
       Tigrinya &\citet{abate_parallel_2018,lakew_low_2020} &75.4k & 30.1k& 2.4k\\
       \bottomrule
    \end{tabular}
    \caption{Benchmark dataset description along with source datasets.}
    \label{tab:dataset}
\end{table}

\subsection{Normalization Settings} \label{sec:normalization}
As discussed in Sec. \ref{sec:background}, there are multiple normalization schemes adopted by prior work, particularly when dealing with Amharic datasets. In this study, we employ three normalization settings: 


\begin{itemize}
    \item \texttt{No-Norm}: We take the dataset as is, without applying any normalization or other alterations. We use this setting as a baseline. 
    \item \texttt{H-only}: We normalize all characters that have the same sound in a given language. We apply this approach for Amharic and Tigrinya, with separate scripts for each language as the characters with the same sound in each language differ (Sec. \ref{sec:background}). We map homophone characters to the most frequent character in the dataset. 
    \item \texttt{HSL}: In this setting, we use the script from \cite{yimam_introducing_2021} and normalize homophone characters, similar sounds, and labialized characters. Since this approach has only been used for Amharic, and there is no standard way to determine ``similar'' sounds, we only apply this approach to the Amharic dataset\footnote{In \cite{yimam_introducing_2021}, characters with `similar' sounds are some characters that have the same consonant but different vowels; for 
    instance, ``\selectlanguage{ethiop}^ci''\selectlanguage{english} /\texttoptiebar{ts} i/ and ``\selectlanguage{ethiop}^ce''\selectlanguage{english} /\texttoptiebar{ts} \textbari/ are mapped to ``\selectlanguage{ethiop}^ce''\selectlanguage{english}. However, there is no standard for determining the similarity of the sounds.}. 

\end{itemize}

In Table \ref{tab:norm_settings}, we give details on how we applied the normalization schemes to our datasets. Note that, for Ge'ez we did not apply any normalization as all characters are distinct--i.e, swapping characters, even if they have the same sound, will result in meaning change (Sec. \ref{sec:background}).

\begin{table}[]
    \centering
    \small
    \begin{tabular}{c|c|c|c}
    \toprule
        \textbf{Language} & \textbf{No Norm} & \textbf{H-Only} & \textbf{HSL} \\
        \midrule
        Amharic  & \checkmark & \checkmark & \checkmark \\
        Tigrinya & \checkmark &\checkmark &- \\
        Ge'ez & \checkmark & - & - \\
        \bottomrule
    \end{tabular}
    \caption{Application of normalization schemes to the three languages of study.}
    \label{tab:norm_settings}
\end{table}


    

    
\section{Experimental Study} \label{sec:experiments}
We investigate the impact of homophone normalization on machine translation(MT) performance. 

In this section, we first give our experimental setup, describing the models we used for our experiments in Sec. \ref{sec:setup}. We conduct experiments on the zero-shot performance of MT systems on sentences with homophone characters (\ref{sec:zero_shot}). We then investigate the impact of normalizing homophone characters in training data for bilingual model training and cross-lingual transfer(\ref{sec:pre_training}). Then, we investigate the efficacy of pots-inference normalization in Sec. \ref{sec:post_inference}. 

\subsection{Experimental Setup} \label{sec:setup}

\subsubsection{Models}

\paragraph{Pre-trained MT Models} For our zero-shot experiments, we used Google Translate\footnote{https://translate.google.com/}, M2M-100-418M \cite{fan_beyond_2021}, and NLLB \cite{nllb_no_2022} models. All three models support Amharic, while Google Translate and NLLB support Tigrinya. However, Ge'ez is not included in any of the three models; hence, we did not perform any zero-shot experiment for English-Ge'ez translation.

\paragraph{Models for Training} To understand the effects of normalization during training, we (1) finetuned the NLLB-600M \cite{nllb_no_2022} model and (2) trained an encoder-decoder Transformer model \cite{vaswani_attention_2017} from scratch. Since the NLLB model includes Amharic and Tigrinya data, we trained the Transformer MT model from scratch to avoid the impact of pre-training in our experimental results. 

\paragraph{Training Details} We trained an encoder-decoder Transformer model with 8 heads and 6 layers. We used an Adam Optimizer~\cite{kingma_adam_2017} with a learning rate of 1e-4 and $\beta1$ =0.9 and $\beta2$=0.98.  We used a learning rate scheduler that decreased the learning rate by a factor of 0.5 if there were no improvements in 2 consecutive epochs. We used Cross Entropy Loss as our loss function and trained the model for 30 epochs. The best model checkpoint based on evaluation set performance was chosen for the final evaluation. 
To fine-tune the NLLB-600M \cite{nllb_no_2022} model, we used the Trainer module from the HuggingFace transformer library~\cite{wolf-etal-2020-transformers}. We fine-tuned the model for 5 epochs with a learning rate of 5e-5 using the default training arguments and a batch size of 32. We used the model's default tokenizer without any additional prefixing. We used the same training scheme for all languages and all experiments.


\subsubsection{Evaluation}
We used both automatic metrics and human evaluation. For automatic metrics, we used BLEU~\cite{papineni-etal-2002-bleu} and ChrF~\cite{popovic-2015-chrf}. BLEU score focuses on overlap in word-level n-grams, whereas ChrF focuses on character-level n-grams. We used the SACREBLEU~\cite{post-2018-call} implementation for both BLEU and ChrF, with their default settings. When calculating the scores, we removed punctuation marks from both reference and prediction sentences. For human evaluation, native speakers of each language qualitatively looked at 50 random sample predictions, comparing the outputs of the different models. We also performed additional error analysis, focusing on words that have characters with the same sound.

\subsection{Zero-Shot Experiments} \label{sec:zero_shot}
This experiment aims to answer \textbf{RQ1}--that is, to understand if there is an existing impact on pre-trained MT models in handling characters with the same sound in the three languages of study.  


\begin{table}[]
    \centering
    \small
    \begin{tabular}{c|c|c|c|c}
    \toprule
     & \multicolumn{2}{c}{\textbf{Amharic}} & \multicolumn{2}{c}{\textbf{Tigrinya}} \\
       Model  & BLEU & ChrF & BLEU & ChrF   \\
        \midrule
       NLLB - 3B  & 10.47 & 34.05 &11.26  & 31.22 \\
       NLLB - 600M & 6.98& 29.16 & 11.55 &31.30  \\
       Google Translate & 9.89 & 33.67 & 16.02 & 38.75 \\
        M2M - 418M & 13.51& 34.78 & - & - \\
       \bottomrule
    \end{tabular}
    \caption{Zero-Shot translation performance.}
    \label{tab:zero_shot}
\end{table}

\paragraph{Results} As can be seen in Table \ref{tab:zero_shot}, all models except the NLLB-200-Distilled-600M have comparable ChrF scores, with M2M-100 having the highest ChrF for Amharic. For Tigrinya, Google Translate had the highest BLEU and ChrF scores. Additionally, M2M-100 has the highest BLEU score for Amharic, while NLLB-200-Distilled-600M had the lowest BLEU score. Further, the open-sourced NLLB-200-Distilled-600M and M2M-100 models performed better than the commercially available Google Translate model for Amharic. 

Qualitatively, we observe that the outputs of the NLLB models for English-Amharic translation usually stick with the Amharic ``Standard'' homophone usage (Sec. \ref{sec:background}). This behavior is not observed in Google Translate. 
For instance, when translating the word ``God,'' all NLLB and M2M models consistently translate it as \selectlanguage{ethiop}'egezi'abe.hEre \selectlanguage{english}, which is consistent with the Ge'ez spelling of the word. However, Google Translate sometimes tends to translate it as \selectlanguage{ethiop}'egezi'abehEre \selectlanguage{english}, switching the ``\selectlanguage{ethiop}.hE''\selectlanguage{english} character with ``\selectlanguage{ethiop}hE''\selectlanguage{english} which does not conform to the standard homophone usage of the word~\cite{aklilu_problems_2010}.

This difference between the open-source models and Google Translate may be due to the fact that open models are trained on publicly available data, which is heavily dominated by religious data for low-resourced languages. 

\begin{table}[]
    \centering
    \small
    \begin{tabular}{p{1.3cm}p{1cm}|c|c|c|c}
    \toprule
   & & \multicolumn{2}{c|}{\textbf{Amharic}}&   \multicolumn{2}{c}{\textbf{Tigrinya}}\\
    \midrule
    Model & Norm. Setting & BLEU & ChrF   & BLEU & ChrF  \\
    \midrule
    \multirow{3}{*}{Transformer}& No Norm  &  \textbf{12.32} & \textbf{29.50}   &\textbf{10.87}  & 25.51  \\
    & H-only & 9.31 & 26.90 & 10.21 &\textbf{26.61} \\
    & HSL & 6.22 & 26.88 & -  & - \\
    \midrule
    
    \multirow{3}{*}{NLLB}& No Norm  &  19.09 & 41.98    &  \textbf{22.71}& \textbf{43.11}   \\
    & H-only & \textbf{19.71} & \textbf{42.59} &  21.41& 41.78 \\
    & HSL & 17.13 & 40.50 & - &- \\
     
    \bottomrule
    \end{tabular}
    \caption{Results for training English-Amharic and English-Tigrinya MT models with different normalization settings (Norm. Setting) applied to training data. Best performance per model type indicated with bold font for each language.}
    \label{tab:monolingual}
\end{table}
\subsection{Effects Training Models with Homophone Normalization} \label{sec:pre_training}
To answer \textbf{RQ2} and \textbf{RQ3}, we trained an encoder-decoder Transformer model from scratch and finetuned an NLLB-600M model as described in Sec. \ref{sec:setup}. We experimented with monolingual and cross-lingual training, which we describe in detail below.

\subsubsection{Monolingual Effects of Normalization} \label{sec:monolingual}
For \textbf{RQ2}, we experimented by training a Transformer model from scratch and finetuning NLLB-600M for each of the three language pairs: Eng-Amh, Eng-Tir, and Eng-Ge'ez. The goal for this experiment was to understand the impact of normalizing homophone characters in the target language during training on the MT performance. As described in Sec. \ref{sec:methods}, we use the \texttt{No-Norm} setting as a baseline for all languages, apply \texttt{H-Only} normalization to Amharic and Tigrinya data, apply \texttt{HSL} normalization to Amharic data only. For Ge'ez, we train without any normalizations (Sec. \ref{sec:background}). 

\paragraph{Results} As can be seen in Table \ref{tab:monolingual}, for both Amharic, the model with \texttt{No-Norm} has better BLEU and ChrF scores as compared to the models trained on normalized data for the Transformer models. For Tigrinya, we observe that the Transformer model has comparable performance with and without normalization. For NLLB-600M, \texttt{H-Only} has a marginal improvement over the \texttt{No-Norm} setting for Amharic (+0.62 BLEU and +0.61 ChrF). The \texttt{HSL} setting has the least performance with NLLB for Amharic. For Tigrinya, we observe that \texttt{No-Norm} has better performance than the \texttt{H-Only} setting.


Qualitatively, we observed that models trained with \texttt{HSL} normalization mostly replace some words with their synonyms and simplify the translation when compared to \texttt{H-Only} and \texttt{No-Norm} settings. Regarding the quality of the translation, NLLB fine-tuned with \texttt{No-Norm} setting provides better translation, preserving the homophone characters in the prediction; this also aligns with the automatic result presented in Table \ref{tab:monolingual}. In the \texttt{H-Only} setting, we noticed that in addition to replacing words with normalized characters, most of the translations were incomplete even though we set the same maximum sequence length for all models.   

\subsubsection{Cross-Lingual Transfer Effects of Normalization}\label{sec:cross_lingual}
For \textbf{RQ3}, we experimented by taking the three models we trained for Amharic as described in Sec \ref{sec:monolingual} and further training with Eng-Tir and Eng-Ge'ez data. In our cross-lingual experiment, the Tigrinya and Ge'ez datasets are taken as is, without any normalization. 

\begin{table*}[]
    \centering
    \small
    \footnotesize
    \begin{tabular}{cc|c|c|c|c|c|c|c|c}

 & \multicolumn{8}{c}{\textbf{Tigrinya}} \\
 \midrule
 & \multicolumn{2}{c|}{No-Transfer} & \multicolumn{2}{c|}{No-Norm} & \multicolumn{2}{c|}{H-only} & \multicolumn{2}{c|}{HSL}\\
 \midrule
Model & BLEU & ChrF&  BLEU & ChrF& BLEU & ChrF& BLEU & ChrF\\   
\midrule
Transformer & 10.87& 25.51& \textbf{12.16} &  \textbf{27.61} & 10.67& 26.24 & 11.23 & 26.44 \\  

NLLB-600M & \textbf{22.71} & \textbf{43.11} &  21.55 & 42.03 & 21.63 & 42.13 & 21.68 & 42.14  \\ 
\midrule

 & \multicolumn{8}{c}{\textbf{Ge'ez}} \\
 \midrule
 & \multicolumn{2}{c|}{No-Transfer} & \multicolumn{2}{c|}{No-Norm} & \multicolumn{2}{c|}{H-only} & \multicolumn{2}{c|}{HSL} \\
 \midrule
Model & BLEU & ChrF&  BLEU & ChrF& BLEU & ChrF& BLEU & ChrF\\   
\midrule
Transformer & 2.46 & 18.72 & \textbf{3.67}  & \textbf{20.80} & 3.56  & \textbf{20.80} & 1.46 & 12.48  \\  

NLLB-600M & 3.36 & 23.48 & 5.22  & 26.54 & \textbf{6.33} & \textbf{28.38} &  6.31 & 28.52 \\ 
    \end{tabular}
    \caption{Performance of MT models in cross-lingual transfer experiments, where \texttt{No-Norm}, \texttt{H-Only}, and \texttt{HSL} refer to models that were initialized with English-Amharic models trained in each of the three settings. The best performance in a row is indicated in bold font. }
    \label{tab:crosslingual}
\end{table*}
\paragraph{Results} As Table \ref{tab:crosslingual} shows, for Tigrinya, we find that the Amharic model that is trained without normalizing the homophone characters--i.e. the \texttt{No-Norm} setting--is a better transfer model as compared to the \texttt{H-Only} and \texttt{HSL} settings when training an MT model from scratch. With finetuning NLLB-200-Distilled-600M, we find that the model directly finetuned from NLLB-200-Distilled-600M performed better than the ones first finetuned on Amharic then finetuned on Tigrinya. With the transfer models for NLLB, we observe comparable performance regardless of the normalization setting with which the Amharic model was finetuned. For Ge'ez, we see that using the Amharic models trained in the \texttt{No-Norm} and \texttt{H-Only} setting provides better BLUE and ChrF scores as compared to using the model trained with \texttt{HSL} setting. Further, we observe that using a Transformer model that was trained with the \texttt{HSL} setting for Amharic has the worst performance when used as a transfer model for Ge'ez. 
Qualitatively, we observe that in both Ge'ez and Tigrinya translations, the output includes code-switching with Amharic words in the translations, changes pronouns, changes gender, and wrongly adds negation of words. 
    
\begin{figure}
    \centering
    \includegraphics[width=\linewidth]{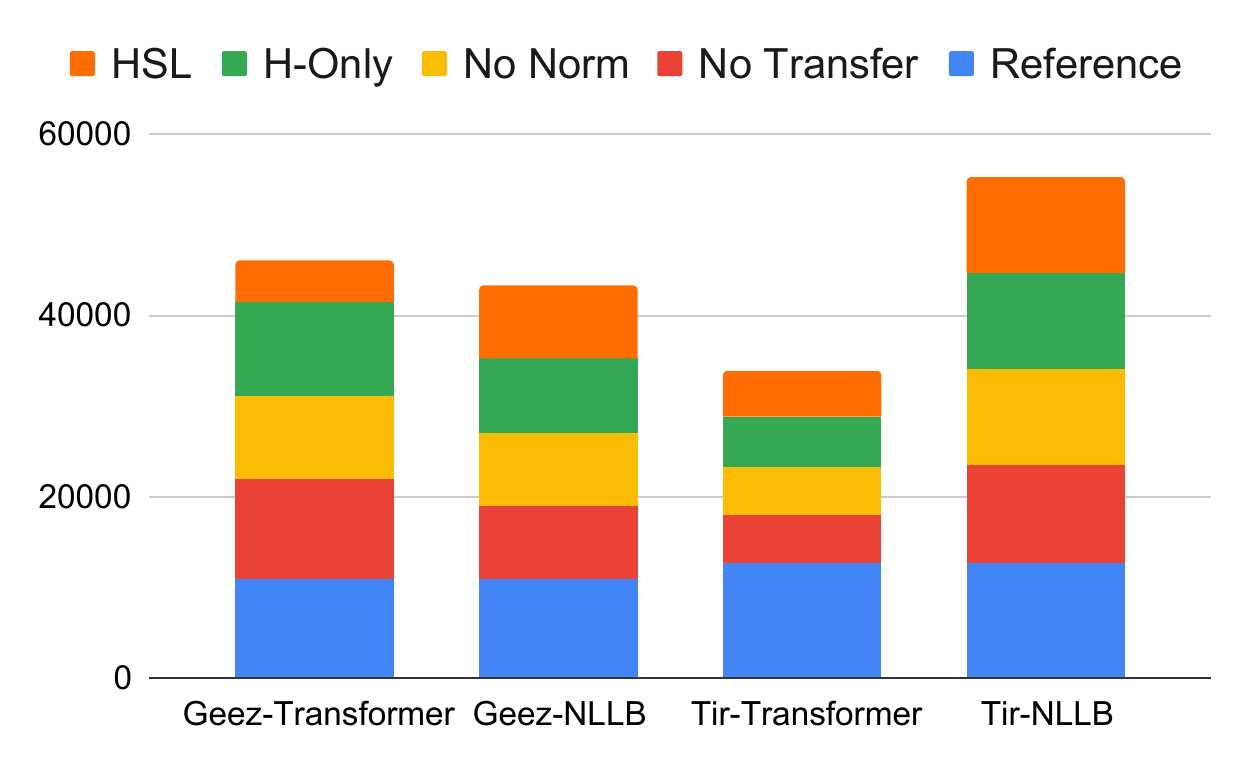}
    
    \caption{Comparison of Unique word count with different transfer settings for English-Tigrinya and English-Ge'ez translation.}
    \label{fig:word_count_comparision}
\end{figure}

We find that the homophone characters that were normalized in the Amharic base model were correctly used in the respective target languages (Tigrinya and Ge'ez). However, the base models trained on the Amharic normalized data tended to repeat characters or words until they reached maximum sequence length instead of translating the source sentence. This is demonstrated in Figure \ref{fig:word_count_comparision}, where the translations with the models trained on Amharic base models with \texttt{HSL} and \texttt{H-Only} settings have fewer unique words in both Ge'ez and Tigrinya, especially for the Transformer model. In Figure \ref{fig:qual_examples}, we provide qualitative examples.

Looking at the character count of the translations in the different transfer settings, we find that all models did not contain a comparable number of characters that were found in the reference dataset; for instance, for Ge'ez the model trained without transfer learning did not contain 33 characters that were in the reference dateset while the model trained with the \texttt{HSL} normalized Amharic base model did not contain 34 characters form the reference. However, training with the Amharic base models added new characters in the predictions, where the characters do not exist in the language. For instance, ``\selectlanguage{ethiop}^sa\selectlanguage{english}'', ``\selectlanguage{ethiop}va\selectlanguage{english}'', ``\selectlanguage{ethiop}^ca''\selectlanguage{english} were added in the Ge'ez predictions although all three characters do not exist in the alphabet for the language.
%
\subsection{Post-Inference Normalization} \label{sec:post_inference}
As discussed in Sec. \ref{sec:background}, normalization of homophones has provided automatic score increases in prior work. However, normalizing the characters before training a model results in models that cannot understand different forms of spelling. Further, as we have seen in Sec. \ref{sec:cross_lingual}, normalizing homophone characters has impacts in transfer learning for languages that use the same writing script. To answer our \textbf{RQ4}, we took the models we trained in the \texttt{No-Norm} setting and applied normalization to the reference and predictions after inference. 
\paragraph{Results}
For the Transformer model we trained, post-normalization improves BLEU and ChrF scores by a small margin (0.24 and 0.29 increase, respectively). For the NLLB finetuned model, we find that applying \texttt{HSL} normalization post-inference boosts the BLEU score by 0.69 and the ChrF by 0.63. From the three models, we find that the NLLB model outperforms the other two on our evaluation dataset. 

We compare how effective post-inference normalization is by including the model from \citet{belay_effect_2022}; we take the model trained without normalization and apply homophone normalization after inference\footnote{We could not compare with the models trained with normalization from \citet{belay_effect_2022} as they are not publicly available.}.
\citet{belay_effect_2022} found a 3.09 BLEU score increase by finetuning an M2M~\cite{fan_beyond_2021} model with \texttt{HSL} normalized data as compared to a model trained with \texttt{No-Norm} data. We cannot directly compare our results with the reported BLEU scores as the test sets are different. 
However, on our evaluation dataset, we find that the model trained by \citet{belay_effect_2022} without applying normalization can have a 1.03 BLEU score increase (Table \ref{tab:post_inference}) with our post-inference scheme. 


    


\begin{table}[]
    \centering
    \small
    \begin{tabular}{c|c|c|c}
    \toprule
      Model  & Setting & BLEU & ChrF \\
      \midrule
      \multirow{3}{*}{\citet{belay_effect_2022}}  & No-Norm & 13.51 & 34.78 \\
        & H-Only & 14.54 & 35.94 \\
         & HSL & \textbf{14.54} & \textbf{35.95} \\
    \midrule
    \multirow{3}{*}{Transformer} & No-Norm & 12.32 & 29.50 \\
        & H-Only & \textbf{12.56} & 29.77 \\
         & HSL & \textbf{12.56} & \textbf{29.79} \\
         \midrule
    \multirow{3}{*}{NLLB-600M} & No-Norm & 19.09 & 41.98 \\
        & H-Only & \textbf{19.78} & 42.60 \\
         & HSL & \textbf{19.78} & \textbf{42.61} \\
         \bottomrule
    \end{tabular}
    \caption{Results for post-inference normalization. Best scores per model indicated with bold font.}
    \label{tab:post_inference}
\end{table}

\section{Discussion}

Our work investigates the impact of homophone normalization for languages that use the Ge'ez script on Machine Translation performance. We provide background on the characteristics of the languages that use the Ge'ez script and detail how prior work used homophone normalization (Sec. \ref{sec:background}). Through a series of experiments (Sec. \ref{sec:experiments}), we demonstrate that homophone normalization does not provide a significant performance gain across all languages, and hurts performance in transfer learning (Sec. \ref{sec:pre_training}). 
As we have discussed in Sec. \ref{sec:background}, homophone normalization has been used as a pre-processing step in the NLP literature for Amharic, setting an implicit standard on what trained models can handle. In this section, we connect this argument to the broader literature on technology-facilitated language change. 

Evolutions in language that are the result of technological constraints make their way to daily lives~\cite{van_dijk_influence_2016}. 
 This is particularly concerning as MT models are used in data creation and augmentation for low-resourced languages~\cite[e.g][]{singh_global_2025}. Machine Translated datasets are also used to train other NLP models~\cite[e.g][]{joshi_adapting_2025}, perpetuating the normalization effect to tasks beyond translation. 

As we have seen in Sec. \ref{sec:experiments}, while normalization has resulted in score improvements in prior work, it affects the performance of models in transfer learning. 
Multiple languages use the same writing script; hence, it is important to consider how the standards we set for one language affect other languages. There might also be dialect differences in how words are spelled, which will not be accounted for when we normalize homophone characters without such considerations.

As the number of low-resourced languages represented in NLP research increases, it is imperative to consider how pre-processing steps applied to these languages alter the overall landscape of language use.
Design decisions could lead to constraints on how and if people can use their language~\cite{wenzel_designing_2024}.
In the context of our study, training models on normalized data results in models that cannot handle alternative spellings. 
For instance, \cite{belay_impacts_2021} found that normalization helped improve performance in information retrieval. However, the performance improvement would require users to conform to the normalized form of spelling. This impact is not limited to homophone normalization; \citet{adebara-abdul-mageed-2022-towards} argue that normalizing tone diacritics, which are essential for lexical disambiguity, affects the usability of retrieval systems for African language speakers. 

 Further, relying solely on automatic score improvements obfuscates the impact of our design decisions beyond its intended effect. 
Instead, our solutions should (\textbf{1}) focus on changing the methods (e.g. the metrics used for evaluation), (\textbf{2}) be explicit under what context the improvements are achieved, and (\textbf{3}) explore alternatives that do not impact the model's ability to handle different versions of a language.
As we propose in Sec. \ref{sec:post_inference}, we can use post-inference interventions to increase automatic scores without altering the training data. While the performance improvement is not as significant as training on normalized data, it is a tradeoff for having a model that can account for different spellings, dialects, and transfer capabilities.


\section{Conclusion} \label{sec:conclusion}
We investigated the impact of homophone normalization on languages that use the Ge'ez script. We find that normalization of homophones in training data leads to poor transfer learning performance for related languages. Furthermore, we find that normalization does not always lead to performance improvement across all languages. We argue against implicit standardization via pre-processing tools and offer an alternative approach that preserves features of the languages during training. We use our work as a case study to call for a more thorough examination of pre-processing steps, particularly for low-resource languages. 

\section*{Limitations}
While our experiments show an increase in the BLEU score with post-inference homophone normalization, we did not conduct a full-scale human evaluation of translation quality; instead, we manually inspected 50 outputs across all normalization settings. Future work should include large-scale human evaluations.  Our conclusion is mainly based on BLEU and chrf++ scores, while they remain standard evaluation tools for MT, they might not show the changes in prediction when we use different normalization settings. 



\bibliography{anthology,references}
\bibliographystyle{acl_natbib}

\appendix

\section{Qualitative Examples}
\label{sec:appendix}

\begin{figure*}[h!]
    \centering
    \includegraphics[width=\linewidth]{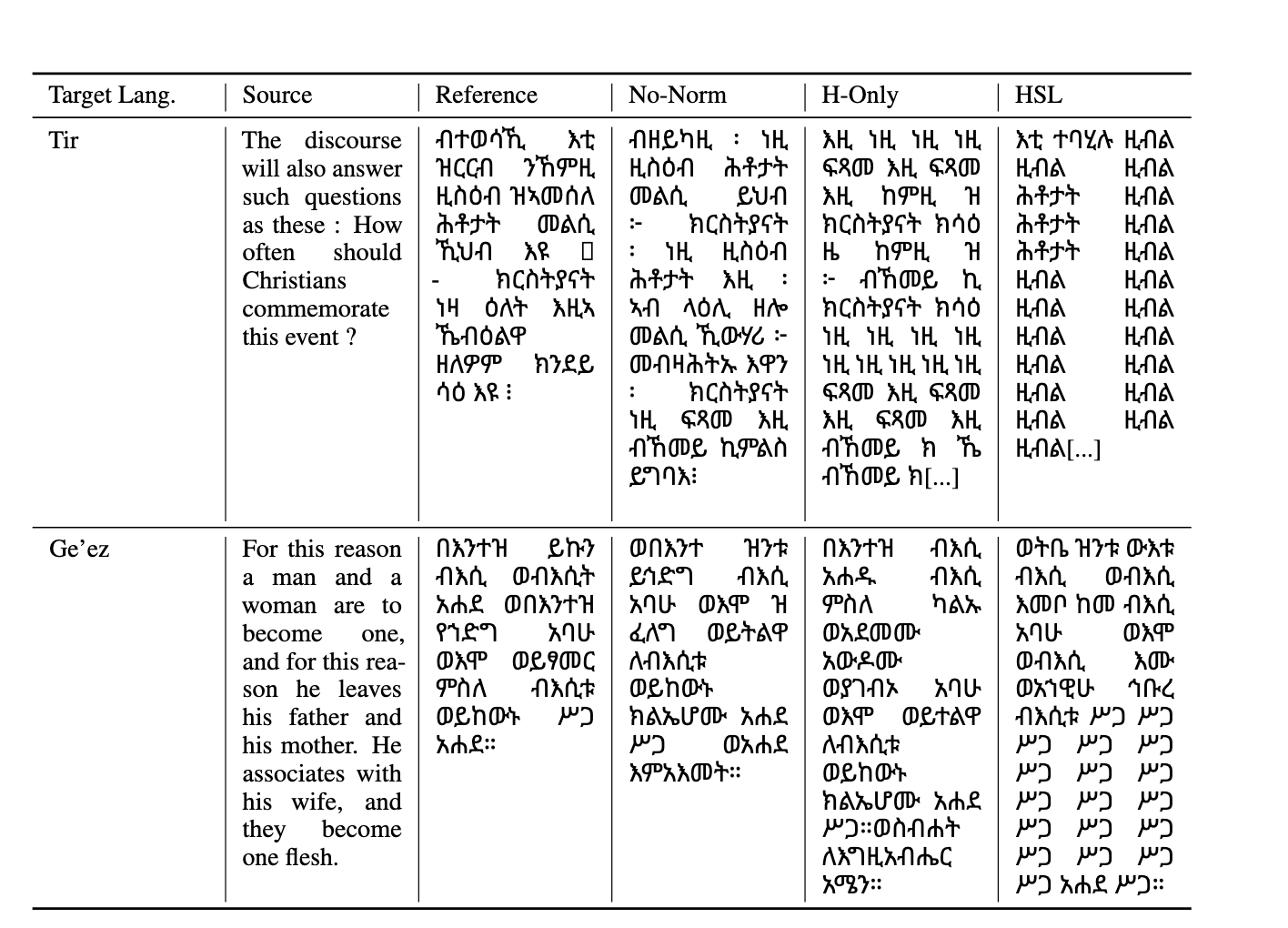}
    \caption{Qualitative examples with transfer learning experiments where the base Amharic model is trained in No-Norm, H-Only and HSL settings}
    \label{fig:qual_examples}
\end{figure*}

In Figure \ref{fig:qual_examples}, we provide qualitative examples for Tigrinya and Ge'ez transfer learning experiments. As the table shows, Amharic models trained with normalization repeat words until they reach maximum sequence length or end of sentence token \selectlanguage{ethiop}(::)\selectlanguage{english}.

\end{document}